\pdfoutput=1

\documentclass[11pt]{article}

\usepackage{EMNLP2022}

\usepackage{times}
\usepackage{latexsym}

\usepackage[T1]{fontenc}

\usepackage[utf8]{inputenc}

\usepackage{microtype}

\usepackage{inconsolata}

\usepackage{hyperref}
\usepackage{url}
\usepackage{microtype}
\usepackage{graphicx}
\usepackage{amsmath}
\usepackage{amsthm}
\usepackage{booktabs}
\usepackage{algorithm}
\usepackage{algorithmic}
\usepackage{wrapfig}
\usepackage{color}
\usepackage{bm}
\usepackage{xr}
\usepackage{multirow}

\usepackage{footnote}

\usepackage{amsmath}
\usepackage{amssymb}

\title{Mask the Correct Tokens: An Embarrassingly Simple Approach \\for Error Correction}

\author{Kai Shen$^{1}$\thanks{\textsuperscript{$\ast$} Both authors contributed equally. This work was conducted when the first and second authors
were interns in Microsoft.}, Yichong Leng$^{2}$\footnotemark[1], Xu Tan$^{3}$\thanks{\textsuperscript{$\ast$} Corresponding author} \\
\textbf{Siliang Tang$^{1}$\footnotemark[2] , Yuan Zhang$^{4}$,  Wenjie Liu$^{4}$, Edward Lin$^{4}$} \\
$^{1}$Zhejiang University, $^{2}$University of Science and Technology of China \\ $^{3}$Microsoft Research Asia, $^{4}$Microsoft Azure Speech \\
$^1$\texttt{\{shenkai, siliang\}@zju.edu.cn}, $^2$\texttt{lyc123go@mail.ustc.edu.cn}\\ 
$^{3,4}$\texttt{\{xuta, yuanz, liwenjie, edlin\}@microsoft.com}}

\begin{document}
\maketitle
\begin{abstract}
Text error correction aims to correct the errors in text sequences such as those typed by humans or generated by speech recognition models.
Previous error correction methods usually take the source (incorrect) sentence as encoder input and generate the target (correct) sentence through the decoder. Since the error rate of the incorrect sentence is usually low (e.g., 10\%), the correction model can only learn to correct on limited error tokens but trivially copy on most tokens (correct tokens), which harms the effective training of error correction. In this paper, we argue that the correct tokens should be better utilized to facilitate effective training and then propose a simple yet effective masking strategy to achieve this goal.
Specifically, we randomly mask out a part of the correct tokens in the source sentence and let the model learn to not only correct the original error tokens but also  predict the masked tokens based on their context information. Our method enjoys several advantages: 1) it alleviates trivial copy; 2) it leverages effective training signals from correct tokens; 3) it is a plug-and-play module and can be applied to different models and tasks. Experiments on spelling error correction and speech recognition error correction on Mandarin datasets and grammar error correction on English datasets with both autoregressive and non-autoregressive generation models show that our method improves the correction
accuracy consistently\footnotemark[1].
\footnotetext[1]{The code for our model is provided for research purposes: \href{https://github.com/microsoft/NeuralSpeech}{https://github.com/microsoft/NeuralSpeech}.}
\end{abstract}

\section{Introduction}
Text error correction techniques have been widely used to correct the errors in text, such as grammar errors~\cite{rothe2021simple,grundkiewicz2019neural,omelianchuk2020gector} or spelling errors~\cite{cheng-etal-2020-spellgcn,zhang-etal-2020-spelling} and automatic speech recognition (ASR) errors~\cite{cucu2013stat,d2016automatic,anantaram2018repairing,linchen2021improve}, etc.
Text error correction is a challenging task because in realistic ASR transcripts or human-typed documents, it is common that the errors are sparse (for example, the error rate of an ASR system is usually below 10\%)~\cite{fastcorrect}. 
To handle the problem, previous works aim to improve the accuracy of correction models by using error detector-corrector architecture and focusing on two directions: 1) promoting the accuracy of error detection~\cite{zhang-etal-2020-spelling}, 2) enhancing the capability of correction~\cite{cheng-etal-2020-spellgcn,fastcorrect,fastcorrect2}. Unfortunately, they all fall in the short that the conventional correction models always copy the correct tokens directly, which is potentially harmful to the learning of the error correction process.

The learning diagram that directly copies the correct tokens can be harmful in two aspects. Firstly, it affects the learning of the correction procedure of the error tokens. Conceptually, the correction diagram can be divided into two main streams: 1) copy the correct token, 2) correct the error token. Intuitively, the former is much easier and contains less language understanding knowledge. The unbalance of trivial copying task (about 90\% tokens) and hard correction task (about 10\% tokens) is potentially harmful to the training of the error correction model. Secondly, it severely underuses valuable training data and consequently lead to low training efficiency. To be more specific, 
since the errors in input text are sparse, most of the tokens in input text are correct and the correction model only learns a trivial copy mapping from input to output, which means that the model can hardly learns knowledge such as language understanding from correct tokens other than copying. To the best of our knowledge, there is no previous work proposed to better leverage the correct tokens.

In this paper, we propose a simple but effective framework MaskCorrect to alleviate the heavy copy phenomenon and improve the utilization of correct tokens in training data. In detail, during the training of the correction model,
we randomly perturb a proportion of correct tokens to a special token \text{<mask>} in the source sentences and train the correction model to predict the correct tokens from \text{<mask>} token. 
With the help of our method, the correction model can benefit from several aspects: 1)
it learns the language understanding knowledge on correct tokens instead of a trivial copy mapping, and thus alleviating the trivial copy phenomenon;
2) since the \text{<mask>} token does not contain explicit clues about its corresponding token, our method will encourage the model to generate output tokens based more on context information, and thus it alleviates the low training efficiency problem.

We conduct a series of experiments on two tasks: 1) Spelling Error Correction and 2) ASR Error Correction on Mandarin datasets with both auto-regressive and non-autoregressive methods. Furthermore, we also conduct experiments on the Grammar Error Correction task on English datasets in Appendix \ref{sec:gec_exp}.  Experimental results show that our MaskCorrect can consistently boost the conventional correction methods in both tasks. Especially, we outperform the state-of-the-art spelling error correction method by a large margin. 

In addition, it is worth noting that the method proposed in this paper is a general framework and can be potentially extended in situations where there exists lots of trivial copy mapping in training data, such as text summarization and question generation.

\section{Related Works}
In this section, we briefly review the spelling/ASR error correction methods and then discuss the copy phenomenon in closely related NLP tasks.

\paragraph{Spelling/ASR Error Correction} 
The main-stream model on spelling error correction leverage a detector-corrector framework~\cite{cheng-etal-2020-spellgcn,zhang-etal-2020-spelling,guo2021global}, which consists of an error detector to detect error tokens and an error corrector to correct the tokens detected as error tokens. A series of works are proposed to improve the correction accuracy with the help of pretraining methods such as BERT~\cite{zhang-etal-2020-spelling} or pronunciation of word~\cite{zhang-etal-2021-correcting}. As for the ASR error correction, the early-stage correction model for ASR leverages the auto-regressive translation model to correct the errors ~\cite{mani2020asr,liao2020improving,wang2020asr,linchen2021improve}. Recently, non-autoregressive correction models are proposed to correct errors in a fast and accurate manner \cite{fastcorrect,fastcorrect2,du2022crossmodal}. 
However, all the methods in both tasks focus on the learning of correcting error tokens, and the models can only learn a trivial copy mapping for tokens without error. Furthermore, they also neglect the low training efficiency because of the heavy copy phenomenon.

\paragraph{The Copy Phenomenon in NLP tasks} The copy phenomenon and the problems it causes have been noticed in the related NLP tasks. In the abstractive text, the copy (pointer) mechanism has been widely integrated to address the issue of frequently copying words from the source context to the desired summary~\cite{see2017get,wu2020biased}. However, a series of works raise potential shorts. \citet{wang2019concept,ji2020cross} finds that merely copying parts of the original text 
tends to plain and uncreative summary. \citet{wei2019regularizing} proposes a novel regularization method to solve the dilemma that traditional copy mechanisms cannot work properly when the target tokens are not consistent with the source tokens.

When it comes to language pretraining, the copy phenomenon attracts lots of attention from researchers. In the original MLM training, it is a common practice~\cite{devlin2019bert,liu2019roberta,he2020deberta,li2022fine} to only mask up to 15\% of original tokens and recover them from the corrupted context while the remaining other tokens are not trained. ELECTRA~\cite{clark2020electra} claims that the MLM training only learns from the 15\% of the tokens, leading to the severe substantial compute cost and low training efficiency. Instead of MLM, they introduce a new pretraining diagram that some tokens are firstly randomly replaced with alternative tokens and the model is trained with the replaced token detection task where it learns to distinguish real input tokens with plausible generated tokens. \citet{wettig2022should} finds that masking more than 15\% of tokens can bring benefits. With a large mask ratio, the model can learn from more cases and thus boosting the training efficiency.

Although the copy phenomenon is still common in the correction, compared to the pretraining which aims to learn the general and robust contextual knowledge, the correction model has a much cleaner purpose that recovering the incorrect tokens mistaken by humans. In this paper, we propose to perturb correct tokens to utilize these correction tokens to help the model learn knowledge about context understanding.

\section{Method}

\subsection{Preliminary Discussion}
\label{sec:preliminary}
In this section, we first conduct some preliminary experiments to qualitatively discuss the impact of the copy phenomenon. In addition, we discuss a simple solution to alleviate the proposed copy problem. Without loss of generality, we conduct preliminary experiments on FastCorrect~\cite{fastcorrect} which is a state of the art (SOTA) baseline in ASR Error Correction (we will introduce the task and FastCorrect in Sec. \ref{sec:asr-exp}).

\paragraph{Add Additional "copy" Tokens.} Based on the assumption that heavy copy is harmful to the correction, one natural question is: what will happen if we make the "copy" more frequently? 

\begin{figure}[h]
  \centering
  \includegraphics[width=1\linewidth]{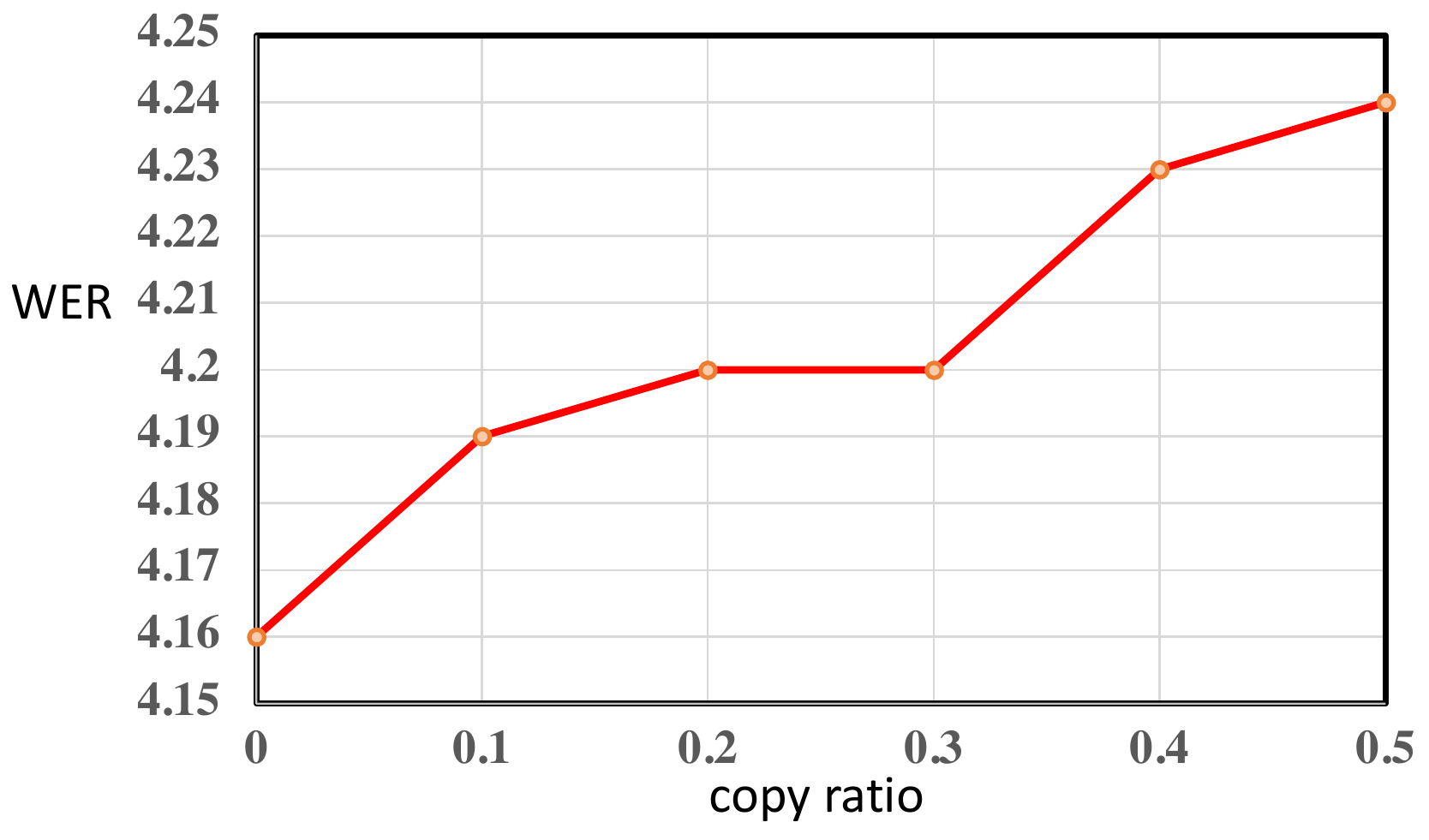}
  \caption{The performance of FastCorrect with different ratios of training "copy" ratio in test set of AISHELL-1. The WER is in \%. Note: the smaller WER is, the better performance the correction model has.}
  \label{fig:exp_asr_redundant}
\end{figure}

To answer this question, we gradually increase the "copy" tokens in the training set while keeping the number of the error tokens unchanged. In detail, we first find out all the samples denoted as $\mathcal{C}$ in the paired training data  whose input sentence \textit{is equal to} the target sentence. Then we randomly select $q$ (denoted as the "copy" ratio $q$) samples from collection $\mathcal{C}$ and add them to the paired training set. Finally, we train the FastCorrect on the synthetic training set. We vary the "copy" ratio $q \in \{0.1, 0.2, 0.3, 0.4, 0.5\}$. We report the word error rate (WER) metric on the AISHELL-1~\cite{bu2017aishell}'s test set in Figure \ref{fig:exp_asr_redundant}.

\begin{figure}[h]
  \centering
  \includegraphics[width=1\linewidth]{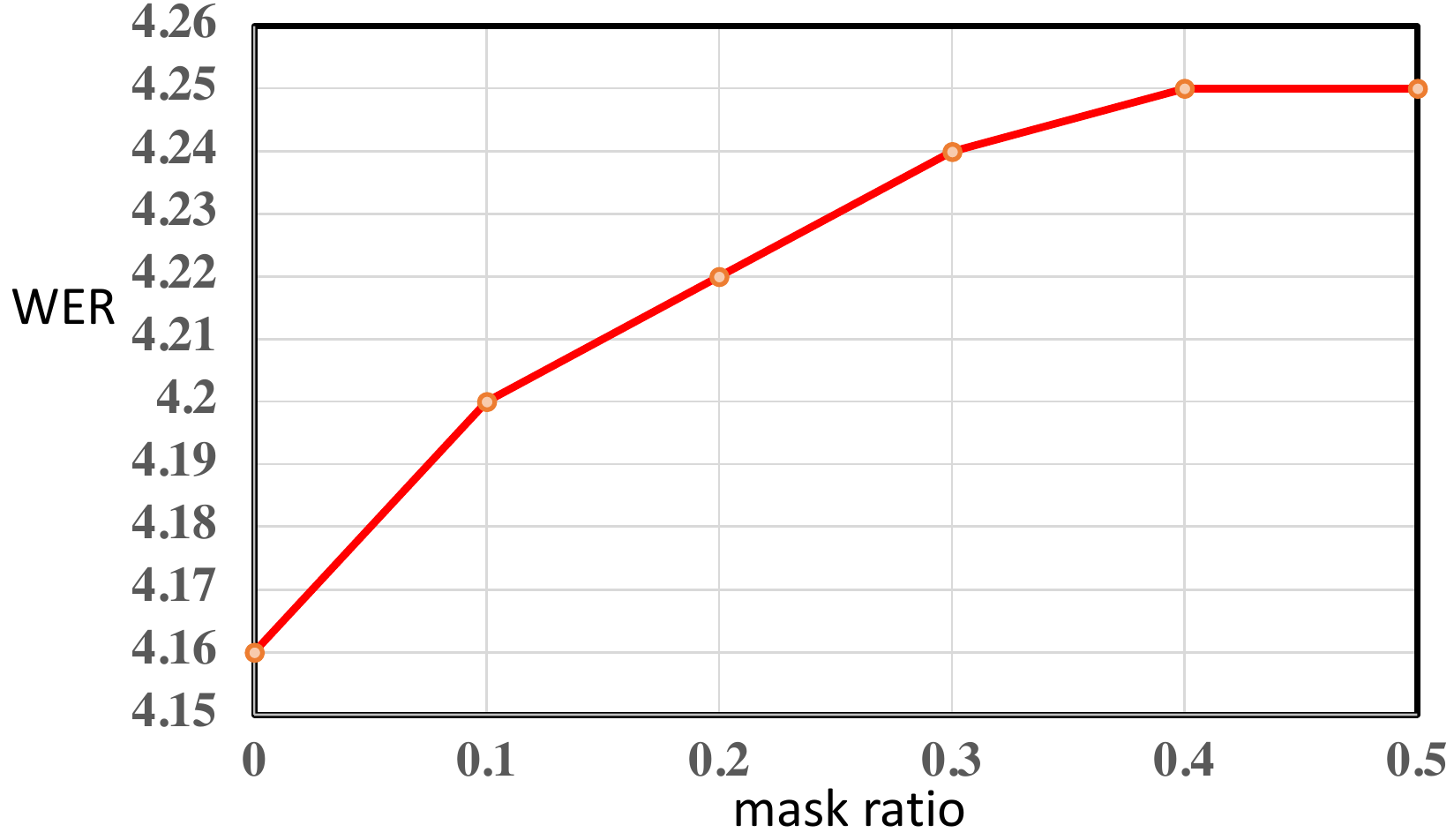}
  \caption{The performance of FastCorrect with different mask ratios of target correct tokens in the test set of AISHELL-1. The WER is in \%. Note: the smaller WER is, the better performance the correction model has.}
  \label{fig:exp_asr_redundant2}
\end{figure}

From the results, we can find that as the "copy" ratio increases, the performance of FastCorrect will decrease gradually. Although FastCorrect is trained with the same amount of error samples, the model is paying more attention to the simply-copied correct samples. This phenomenon illustrates that the heavy copy phenomenon is harmful to the correction, which proves the correctness of our motivation. 
\paragraph{Mask on Target Loss.} 
To alleviate the heavy copy problem, one direct but natural idea is to mask off the gradients of some correct tokens during the gradient back-propagation. We implement this idea by simply masking off a specific number of correct tokens in the loss calculation. The results are shown in Figure \ref{fig:exp_asr_redundant2}.

From the results, surprisingly, we can find that the WER results increase when we gradually mask more correct tokens. This experiment indicates that although the correct tokens are copied as a phenomenon, they cannot be easily dropped during the loss calculation. We have several insights: 1) we cannot alleviate the "copy" by preventing the loss in back-propagation. Every token plays an important role in the context and the overall distribution will be influenced if we simply prevent back-propagating gradient of some tokens; 2) Although the correct tokens are copied as a result, they still contain information in some way and cannot be ignored easily.

\begin{figure*}[ht]
  \centering
  \includegraphics[width=0.85\linewidth]{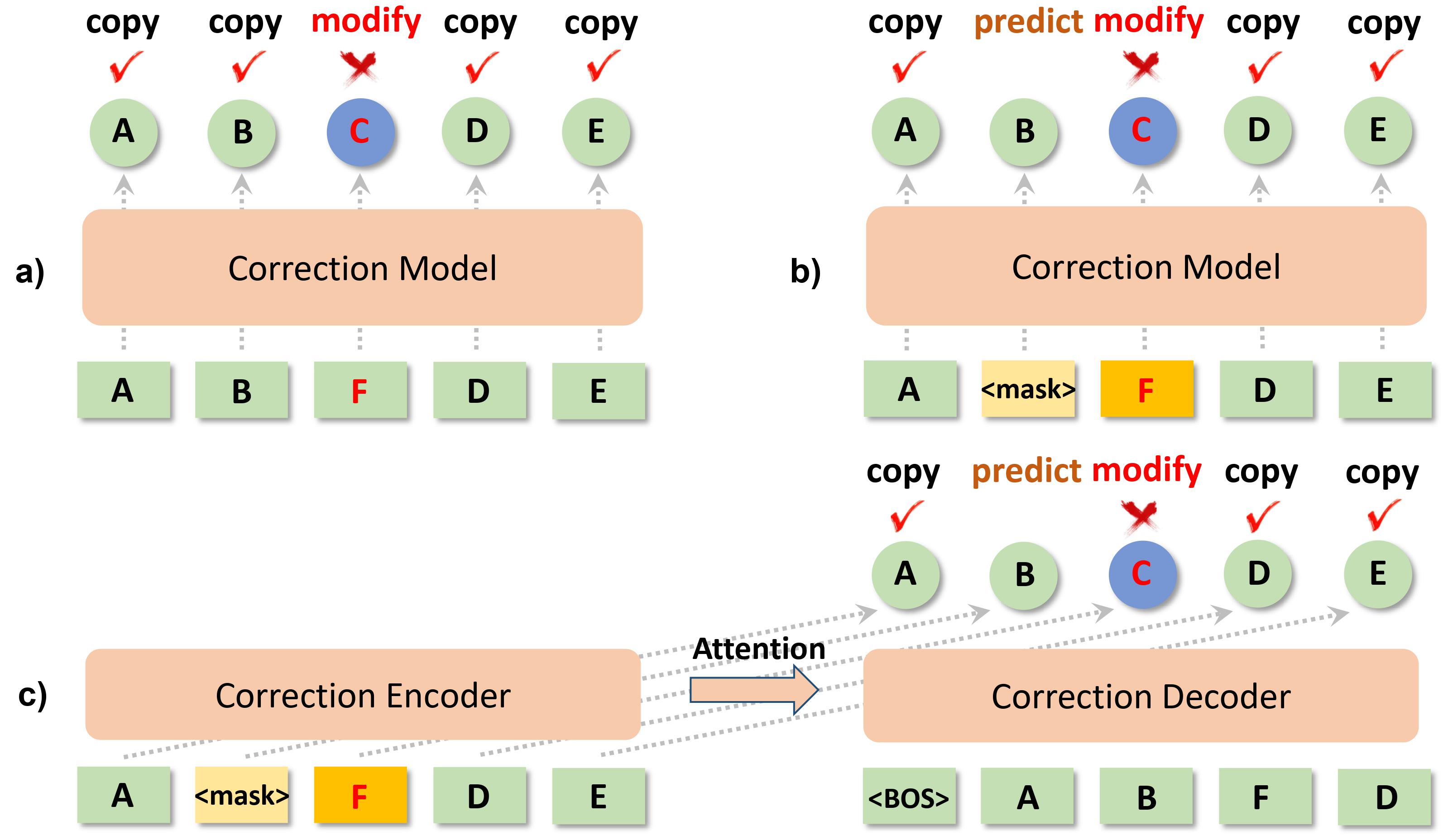}
  \caption{The conventional correction framework (a) and our proposed MaskCorrect in both non-autoregressive (b) and autoregressive (c) framework.}
  \label{fig:framework}
\end{figure*}

Motivated by this experiment, in the MaskCorrect framework, we choose to mask the correct tokens in the source sentence and predict the corresponding token. There are several advantages: 1) every token participates in the correction loss back-propagation, which does not affect the overall distribution; 2) by masking off the correct tokens in the source sentence, the model is encouraged to predict the token only by the context.

\subsection{MaskCorrect}

In this subsection, we introduce our method for addressing the heavy copy of correct tokens in the training data of correction models.

As shown in Figure \ref{fig:framework} (a), the conventional correction models focus on the correction of error tokens (i.e., token ``F'' in Figure \ref{fig:framework}) and simply learn a copy mapping on the rest of the tokens that have no error (i.e, token ``A'', ``B'', ``D'' and ``E'' in Figure \ref{fig:framework}). If the effective training samples in training data are the tokens with error only, then the valuable training data is heavily underused since only a small proportion of tokens in training data (typically about 10\%) contains errors.

We encourage the model to leverage correct tokens by introducing perturbation, as shown in Figure \ref{fig:framework} (b) in non-autoregressive and (c) in auto-regressive way. Specifically, we random perturb $p$\% input tokens to a special token \text{<mask>} before the input tokens are fed into the correction model and keep the target tokens untouched. As a result, the model needs to predict the masked token based on its context information (i.e., predicting token ``B'' based on its context information in Figure 1 (b) and (c)). A better understanding of context information will also benefit the correction of error tokens since the model needs to leverage context information to detect which token has an error and how to correct this error token.

We would like to highlight that the perturbation is only conducted on the input tokens, and thus our method can be used on any error correction tasks with arbitrary model architecture or loss function designed for certain error correction tasks. Our  perturbation helps the model better utilize correction tokens by a mask-and-predict task.

\section{Experiments}

In this section, we evaluate our method on two tasks, namely \textit{ASR Error Correction} and \textit{Spelling Error Correction} on Mandarin datasets. We first report the accuracy of the baselines and our methods to demonstrate the effectiveness of our method on each task. Then we conduct ablation experiments to comprehensively analyze the effect of the mask mechanism. Furthermore, we also conduct experiments on \textit{Grammar Error Correction} on English corpus in Appendix \ref{sec:gec_exp} to demonstrate the effectiveness.

\subsection{Spelling Error Correction}

\begin{table*}[]
\centering
\begin{tabular}{c|l|rrrr|rrrr}
\toprule
\multirow{2}{*}{Test Sets} & \multicolumn{1}{c|}{\multirow{2}{*}{Model}} & \multicolumn{4}{c|}{Detection Level}                                                            & \multicolumn{4}{c}{Correction Level}                                                           \\ \cline{3-10} 
                           & \multicolumn{1}{c|}{}                       & \multicolumn{1}{c}{Acc.} & \multicolumn{1}{c}{P.} & \multicolumn{1}{c}{R.} & \multicolumn{1}{c|}{F1} & \multicolumn{1}{c}{Acc.} & \multicolumn{1}{c}{P.} & \multicolumn{1}{c}{R.} & \multicolumn{1}{c}{F1} \\ \midrule
\multirow{6}{*}{SIGHAN14}  & FASPell                                     & \multicolumn{1}{c}{-}  & 61.00               & 53.50               & 57.00                 & \multicolumn{1}{c}{-}  & 59.40               & 52.00               & 55.40                \\ \cline{2-10} 
                           & SpellGCN                                    & 69.83               & 69.20               & 65.20               & 67.14                 & 68.92               & 70.10               & 60.94               & 65.20                \\ \cline{2-10} 
                           & SoftMask                                    & 69.49               & 72.69               & 60.38               & 65.97                 & 68.36               & 71.90               & 58.08               & 64.26                \\
                           & + MaskCorrect                               & 71.37               & 74.11               & 63.85               & 68.60                 & 70.72               & 73.70               & 62.50               & 67.64                \\ \cline{2-10} 
                           & BERT                                        & 68.27               & 69.10               & 63.65               & 66.27                 & 66.85               & 68.10               & 60.77               & 64.23                \\
                           & + MaskCorrect                               & \textbf{71.94}               & \textbf{73.82}               & \textbf{66.15}               & \textbf{69.78}                 & \textbf{70.90}               & \textbf{73.19}               & \textbf{64.04}               & \textbf{68.31}                \\ \midrule \midrule
\multirow{6}{*}{SIGHAN15}  & FASPell                                     & \multicolumn{1}{c}{-}  & 67.60               & 60.00               & 63.50                 & \multicolumn{1}{c}{-}  & 66.60               & 59.10               & 62.60                \\ \cline{2-10} 
                           & SpellGCN                                    & 78.92               & 80.80               & 75.70               & 78.17                 & 78.10               & 80.10               & 72.70               & 76.22                \\ \cline{2-10} 
                           & SoftMask                                    & 77.36               & 79.96               & 72.14               & 75.85                 & 76.45               & 79.54               & 70.30               & 74.63                \\
                           & + MaskCorrect                               & 78.45               & 80.32               & 74.54               & 77.32                 & 77.09               & 79.71               & 71.77               & 75.53                \\ \cline{2-10} 
                           & BERT                                        & 78.82               & \textbf{82.12}               & 72.88               & 77.22                 & 77.64               & \textbf{81.62}               & 70.48               & 75.64                \\
                           & + MaskCorrect                               & \textbf{80.91}               & 81.56               & \textbf{79.15}               & \textbf{80.34}                 & \textbf{79.00}               & 80.79               & \textbf{75.28}               & \textbf{77.94}                \\ \bottomrule
\end{tabular}
\caption{The performance of the baselines and our method for Spelling Error Correction. We report the accuracy (Acc.), precision (P.), recall (R.) and F1 scores. We report the FASPell's performance in \cite{cheng-etal-2020-spellgcn}. Best results are in bold. All metrics are in \%.}
\label{table:spelling_main_results}
\end{table*}

In this section, we evaluate the effectiveness of our method on spelling error correction task.
\subsubsection{Datasets}

We conduct the experiments on two publicly available SIGHAN datasets: SIGHAN 2014~\cite{yu2014chinese} (SIGHAN14) and SIGHAN 2015~\cite{tseng2015introduction} (SIGHAN15) from Chinese Spell Check Bake-offs. Following \citet{cheng-etal-2020-spellgcn,guo2021global}, we adopt the standard split of the training and testing data of SIGHAN, as well as the same data preprocessing which converts the traditional Chinese to simple Chinese by OpenCC.

From the previous works~\cite{cheng-etal-2020-spellgcn,guo2021global}, the SIGHAN is too small to achieve acceptable results, we follow their common practice and collect 5 million unpaired texts from the internet to pretrain the model. The detailed procedure to add noises can be found in Appendix \ref{appendix:add-noise_spelling}.
For the finetune split, we include the training corpus of SIGHAN14 and SIGHAN15. In addition, we also include the 271K samples generated by automatic method~\cite{wang-etal-2018-hybrid} as the finetune data\footnotemark[2]. The detailed data statistics are shown in Appendix \ref{appendix:table-spelling_dataset_statistics}.
\footnotetext[1]{\href{http://ir.itc.ntnu.edu.tw/lre/sighan7csc.html}{http://ir.itc.ntnu.edu.tw/lre/sighan7csc.html}}
\footnotetext[2]{\href{https://github.com/wdimmy/Automatic-Corpus-Generation/blob/master/corpus/train.sgml}{https://github.com/wdimmy/Automatic-Corpus-Generation/blob/master/corpus/train.sgml}}

We use the pretrained Bert tokenizer provided by HuggingFace~\cite{wolf-etal-2020-transformers} to tokenize the sentences. In addition, we filter out the sentences whose length are larger than 100 for the paired and unpaired training corpus.

\subsubsection{Baseline Methods}

\paragraph{FASPell} The FASPell~\cite{hong-etal-2019-faspell} proposes a new correction paradigm that consists of a denoising autoencoder and a decoder. Instead of using an inflexible and insufficient confusion dictionary, it utilizes the similarity metric to select the candidates in the decoder.

\paragraph{SpellGCN} the SpellGCN~\cite{cheng-etal-2020-spellgcn} incorporates the phonological and visual similarity knowledge into language modeling. They propose to build the similarity graph for each word token in the view of the pronunciation and word shape.

\paragraph{BERT} The BERT~\cite{devlin2018bert} is a commonly used baseline for spelling error correction. It simply employs the pretrained BERT as a sentence encoder network. We take the encoded sentence features and feed them to  network with multiple linear layers and ReLU activation to predict the targets.

\paragraph{SoftMask BERT} This method~\cite{zhang-etal-2020-spelling} is a BERT-based spelling error correction baseline, which consists of a GRU-based error detection module and a BERT-based error correction module. A soft-masking gate softly fuses the \text{<mask>} token's embedding with the initial word embedding by the probability of being an error token predicted by the detection module. Then the fused embeddings are 
fed to the correction module.

\begin{table*}[t]
\begin{tabular}{l|r|rrrr|rrrr}
\toprule
                          & \multicolumn{1}{l|}{}           & \multicolumn{4}{c|}{Detection Level}                                                                                 & \multicolumn{4}{c}{Correction Level}                                                                                \\ \hline
                          & \multicolumn{1}{c|}{Mask Ratio} & \multicolumn{1}{c}{Acc.(\%)} & \multicolumn{1}{c}{P.(\%)} & \multicolumn{1}{c}{R.(\%)} & \multicolumn{1}{c|}{F1(\%)} & \multicolumn{1}{c}{Acc.(\%)} & \multicolumn{1}{c}{P.(\%)} & \multicolumn{1}{c}{R.(\%)} & \multicolumn{1}{c}{F1(\%)} \\ \midrule
\multirow{7}{*}{SIGHAN14} & 0\%                            & 68.27                        & 69.10                      & 63.65                      & 66.27                       & 66.85                        & 68.10                      & 60.77                      & 64.23                      \\ \cline{2-10} 
& 10\%                            & 65.82                        & 66.05                      & 64.23                      & 65.49                       & 65.82                        & 66.05                      & 62.12                      & 64.02                      \\ \cline{2-10} 
                          & 15\%                            & 71.85                        & 73.87                      & 65.77                      & 69.58                       & 70.80                        & 73.29                      & 63.85                      & 68.24                      \\ \cline{2-10} 
                          & \textbf{20\%}                            & \textbf{71.94}                        & 73.82                      & \textbf{66.15}                      & \textbf{69.78}                       & 70.90                        & 73.19                      & \textbf{64.04}                      & \textbf{68.31}                      \\ \cline{2-10} 
                          & 25\%                            & 72.60                        & \textbf{76.08}                      & 64.23                      & 69.66                       & \textbf{71.56}                        & \textbf{75.47}                      & 62.12                      & 68.14                      \\ \cline{2-10} 
                          & 30\%                            & 68.36                        & 68.85                      & 64.62                      & 66.67                       & 67.80                        & 68.46                      & 63.46                      & 65.87                      \\ \cline{2-10} 
                          & 40\%                            & 66.29                        & 66.01                      & 64.23                      & 65.11                       & 64.78                        & 64.90                      & 61.15                      & 62.97                      \\ \cline{2-10} 
                          & 50\%                            & 67.61                        & 68.26                      & 63.27                      & 65.67                       & 66.67                        & 67.58                      & 61.35                      & 64.31                      \\ \midrule \midrule
\multirow{7}{*}{SIGHAN15} & 0\%                            & 78.82                       & \textbf{82.12}                      & 72.88                     & 77.22                       & 77.64                        & \textbf{81.62}                      & 70.48                      & 75.64                      \\ \cline{2-10} 
& 10\%                            & 76.64                        & 76.84                      & 75.28                      & 76.05                       & 75.27                        & 76.16                      & 72.51                      & 74.29                      \\ \cline{2-10} 
                          & 15\%                            & 78.27                        & 78.32                      & 77.31                      & 77.81                       & 77.36                        & 77.90                      & 75.46                      & 76.66                      \\ \cline{2-10} 
                          & \textbf{20\%}                            & \textbf{80.91}                        & 81.56                      & \textbf{79.15}                      & \textbf{80.34}                       & \textbf{79.00}                        & 80.79                      & 75.28                      & \textbf{77.94}                      \\ \cline{2-10} 
                          & 25\%                            & 78.36                        & 79.92                      & 74.91                      & 77.33                       & 77.09                        & 79.35                      & 72.32                      & 75.68                      \\ \cline{2-10} 
                          & 30\%                            & 76.36                        & 75.73                      & 76.57                      & 76.15                       & 74.82                        & 74.95                      & 73.43                      & 74.18                      \\ \cline{2-10} 
                          & 40\%                            & 77.82                        & 76.90                      & 78.60                      & 77.74                       & 76.55                        & 76.30                      & \textbf{76.01}                      & 76.16                      \\ \cline{2-10} 
                          & 50\%                            & 76.82                        & 77.54                      & 74.54                      & 76.01                       & 75.18                        & 76.74                      & 71.22                      & 73.88                      \\ \bottomrule
\end{tabular}
\caption{The ablation experiments on the mask ratio hyper-parameter $p$. The best metrics are in bold. All the metrics are in \% and Acc.-Accuracy, P.-Precision, R.-Recall.}
\label{table:spelling_ablation_maskratio}
\end{table*}

\subsubsection{Evaluation Metrics}
We report the sentence level accuracy, precision, recall, and F1 scores as the evaluation metrics, which are commonly used in the spelling error correction tasks. All the metrics are computed by the official evaluation tools\footnotemark[1].
\footnotetext[1]{\href{http://nlp.ee.ncu.edu.tw/resource/csc.html}{http://nlp.ee.ncu.edu.tw/resource/csc.html}}

\subsubsection{Implementation Details}
\label{sec:implement-spelling}
We choose the \textit{BERT} and \textit{SoftMask BERT} as the correction backbone and evaluate the effectiveness of our method. The detailed training configuration can be found in Appendix \ref{appendix:training_config_spell}. For both BERT and SoftMask BERT, given the input sentence with $m$ tokens, we first find out all the correct tokens by comparing the input sentence with the ground truth sentence. Secondly, we randomly select the tokens from the correct token set with ratio $p$ without replacement. For each selected token, it will be replaced with \text{<mask>} token with a probability of $m$, or will be replaced with a randomly chosen other token with a probability of $n$, otherwise it will not be changed with a probability of $1 - m -n$. We apply the MaskCorrect on both unpaired and paired training corpus with the same mask hyper-parameter $p, m, n$. However, since it is inevitable that there is no such \text{<mask>} token during the inference stage, we have to finetune model with the paired data additionally for 3 epochs to cope with the training-inference mismatch problem to achieve better performance.

\subsubsection{Results Analysis}
In this section, we compare our method with the baseline models on the two datasets for Spelling Error Correction. We apply the proposed MaskCorrect method on two baselines, namely \textit{BERT + MaskCorrect} and \textit{SoftMask + MaskCorrect}. The results are shown in Table \ref{table:spelling_main_results}.

First, we compare all the methods and we find the variant \textit{BERT + MaskCorrect} outperforms all other baselines in both SIGHAN14 and SIGHAN15 datasets. 
Specifically, we compare two baselines \textit{BERT} and \textit{SoftMask} with the corresponding MaskCorrect variants. With the MaskCorrect, 1) the \textit{SoftMask} obtains 2.63\% and 3.02\% F1 score gains on detection and correction levels in SIGHAN14 and obtains 1.47\% and 0.9\% gains on SIGHAN15, respectively; 2) the \textit{BERT} obtains 3.51\% and 4.07\% F1 gains and 3.12\% and 2.30\%, respectively. It demonstrates that by randomly perturb a certain proportion of correct tokens into \text{<mask>} tokens, the correction models can learn the semantic knowledge from the correct tokens, which enhances the ability of detection and correction the spelling errors.

\subsubsection{Ablation Study}

In this section, we further study the impact of the masking ratio (i.e., the hyper-parameter $p$). Without loss of generality, we choose the best baseline \textit{BERT + MaskCorrect} as the framework and conduct experiments on both SIGHAN14 and SIGHAN15 with $p$ varying in $\{0.1, 0.15, 0.2, 0.25, 0.3, 0.4, 0.5\}$. The results are shown in Table \ref{table:spelling_ablation_maskratio}.

From the results, we have several observations: 1) when the mask ratio is small, there are no gains or even a slight performance drop. Because when the mask ratio is small, there is no enough knowledge learned from the correct tokens and the mask tokens may hinder the correction of error tokens. 2) when the mask ratio is large, the MaskCorrect cannot bring the expected benefits. Empirically, we find that under a large mask ratio, the cross-entropy loss is always large and cannot be well-optimized even with many epochs. This is because when we mask too much correct tokens, the language information loses too much and the language model cannot learn the correct knowledge from it. 3) the best choice of mask ratio is 0.2 in this task for the BERT correction model.

\subsection{ASR Error Correction} 
\label{sec:asr-exp}
In this section, we evaluate the effectiveness of our method on the ASR Error Correction task.

\begin{table*}[t]
\centering
\begin{tabular}{l|rrrr|rr}
\hline
\multicolumn{1}{c|}{\multirow{3}{*}{Methods}} & \multicolumn{4}{c|}{AISHELL-1}                                                                             & \multicolumn{2}{c}{Inhouse Corpus}                 \\ \cline{2-7} 
\multicolumn{1}{c|}{}                         & \multicolumn{2}{c|}{Test Set}                        & \multicolumn{2}{c|}{Dev Set}                        & \multicolumn{2}{c}{Test Set}                       \\ \cline{2-7} 
\multicolumn{1}{c|}{}                         & \multicolumn{1}{c}{WER $\downarrow$} & \multicolumn{1}{c|}{WERR$\uparrow$}  & \multicolumn{1}{c}{WER$\downarrow$} & \multicolumn{1}{c|}{WERR$\uparrow$} & \multicolumn{1}{c}{WER$\downarrow$} & \multicolumn{1}{c}{WERR$\uparrow$} \\ \hline
No Correction                                 & 4.83                    & \multicolumn{1}{r|}{-}      & 4.46                    & \multicolumn{1}{r|}{-}     & 8.99                    & \multicolumn{1}{r}{-}     \\ \hline
\multicolumn{7}{l}{\textit{Non-autoregressive correction models}}     \\ \hline
LevT(MIter=1)                               & 4.73                    & \multicolumn{1}{r|}{2.07}  & 4.37                    & 2.02                      & 8.86                    & 1.45                     \\
LevT(MIter=3)                                 & 4.74                    & \multicolumn{1}{r|}{1.86}  & 4.38                    & 1.79                      & 8.89                    & 1.11                     \\
FELIX                                         & 4.63                    & \multicolumn{1}{r|}{4.14}  & 4.26                    & 4.48                      & 8.74                    & 2.78                     \\ \hline
FastCorrect                                   & 4.16                    & \multicolumn{1}{r|}{13.87} & 3.89                    & 13.30                     & 8.44                    & 6.12                     \\
+ MaskCorrect                                 & \textbf{4.10}                    & \multicolumn{1}{r|}{\textbf{15.11}} & \textbf{3.82}                    & \textbf{14.35}                     & \textbf{8.35}                    & \textbf{7.12}                     \\ \hline
\multicolumn{7}{l}{\textit{Autoregressive correction models}}     \\ \hline
AR  Transformer                                          & 4.08                    & \multicolumn{1}{r|}{15.53} & 3.80                    & 14.80                     & 8.32                    & 7.45                     \\
+ MaskCorrect                                 & \textbf{4.03}                    & \multicolumn{1}{r|}{\textbf{16.56}} &\textbf{3.75}                    & \textbf{15.92}                   & \textbf{8.25}                    & \textbf{8.23}                  \\ \hline
\end{tabular}
\caption{The performance of the baseline models and our method for ASR Error Correction. Note that we only apply our method on the strongest non-autoregressive correction model. We report the WER and the WERR as the evaluation metrics. All metrics are in \%. "MIter" is a hyper-parameter in LevT controlling max decoding iteration.}
\label{table:asr_main_results}
\end{table*}

\subsubsection{Dataset and ASR Model}
\label{sec:dataset}
We conduct the experiments on two Chinese datasets, the public dataset AISHELL-1~\cite{bu2017aishell} and the inhouse dataset. Following the previous works~\cite{linchen2021improve}, we employ the ASR model to train on the training set. Then the ASR model is used to transcribe the ASR training split again to generate the transcripts. The transcripts along with the corresponding ground truth texts are used to train the ASR correction model. The validation and test splits for ASR correction are inferred by the above ASR model. The details of dataset statistics and ASR model can be found in Appendix \ref{appendix:table-asr_dataset_statistics} and \ref{appendix:asr_model_details}, respectively.

However, for AISHELL-1 dataset, it's too small to achieve good correction performance~\cite{fastcorrect,fastcorrect2}. To address this issue, we follow the previous works'~\cite{fastcorrect,fastcorrect2} practice to construct a much larger pseudo training corpus to pretrain the correction model. Details can be found in Appendix \ref{appendix:add-noise_asr}. As for the inhouse dataset, since it's large enough to achieve good results, we don't apply unpaired data pretraining and directly train the correction model from scratch.

\subsubsection{Baseline Methods and Evaluation Metric}

\paragraph{FELIX} FELIX~\cite{mallinson2020felix} is a NAR algorithm for text edition. It aligns the mismatch tokens between the source and target sentences by two sub-tasks: 
1) decide the subset of tokens in source sentences and their order in the target sentences by tagging operation 
and 2) insert tokens to infill the missing tokens in the target sentences.

\paragraph{LevT} Levenshtein Transformer (LevT)~\cite{gu2019levenshtein} is a partially autoregressive seq2seq model devised for flexible sequence generation tasks. It implicitly predicts the insertion and deletion with multiple iterations in the decoder. 

\paragraph{Autoregressive Correction Model (AR Transformer)} For the autoregressive sequence generation model, we follow the standard Transformer~\cite{vaswani2017attention} which takes the sentences with potential errors as input and outputs the corrected sentences autoregressively. 

\paragraph{FastCorrect} FastCorrect~\cite{fastcorrect} is a non-autoregressive ASR correction model based on edit distance. It employs edit distance to find the fine-grained mapping between error tokens and the corresponding correct tokens based on edit operation: insertion, deletion and substitution. We follow the default setting and implement it based on the official code\footnotemark[1].
\footnotetext[1]{\href{https://github.com/microsoft/NeuralSpeech/tree/master/FastCorrect}{https://github.com/microsoft/NeuralSpeech/tree/master\\/FastCorrect}}

As for the evaluation metrics, we follow the practice in previous works and adopt the word error rate (WER) and Word Error Reduction Rate (WERR) as the automatic evaluation metric. We use the official code in ESPnet to calculate it.
    
\subsubsection{Implementation Details}
\label{sec:mask-implement}
We evaluate our method on two SOTA ASR correction model in both AR and NAR, namely \textit{AR Transformer} and \textit{FastCorrect}. Similar to the Spelling Error Correction~\ref{sec:implement-spelling}, we first collect the correct tokens from inputs and randomly mask the tokens with ratio $p$ and probability $m$ and $n$. It is worth noting that we apply the MaskCorrect on both unpaired and paired training corpus for AISHELL-1 and only paired training corpus for inhouse dataset (since we do not use unpaired training corpus for the inhouse dataset as mentioned in Section \ref{sec:dataset}). In addition, since there is no \text{<mask>} token during inference, we have to finetune on the original data without mask for 3 epochs. Please refer to Appendix \ref{appendix:training_config_asr} for more details.

\subsubsection{Results Analysis}

In this section, we compare our method with the baseline models on the two datasets for ASR Error Correction. We apply the proposed MaskCorrect framework on two baselines: \textit{FastCorrect} and \textit{AR Transformer}, for NAR and AR correction diagrams, respectively. Note that we only apply our method to the strongest non-autoregressive correction model, FastCorrect. The results are shown in Table \ref{table:asr_main_results}.

From the table, we can have the following observations. Firstly, the \textit{AR + MaskCorrect} model achieves the best performance compared with all other baselines, which demonstrates the effectiveness of our method. Secondly, by comparing the FastCorrect and AR Transformer with the corresponding variants, we can find that the MaskCorrect can enhance both baselines on all datasets. Concretely, with the MaskCorrect, the FastCorrect can further reduce the WER (measured by WERR) of ASR model by 1.24\% and 1\% on the test sets of the AISHELL-1 and inhouse dataset, respectively. Meanwhile the AR Transformer can reduce further the WER by 1.03\% and 0.78\%, respectively. It confirms that the MaskCorrect is useful for both AR and NAR methods. By adding a perturb (i.e., masking off some correct tokens), the correction model can learn from the correct tokens, which is beneficial to the ASR error correction.

\subsubsection{Ablation Study}

In this section, we conduct an ablation study to verify the affect of the mask ratio $p$. We choose the best NAR baseline FastCorrect + MaskCorrect as the framework and apply different value of $p$ varying in $\{0.1, 0.15, 0.2, 0.25, 0.3, 0.4, 0.5\}$ on the training set. The results are shown in Table \ref{table:asr_ablation_mask_ratio}.

From the results, we have several observations: 1) when the mask ratio falls in [0, 0.2], the MaskCorrect will bring a positive impact to the FastCorrect, otherwise it will cause a slight performance drop. We think when the mask ratio is too large, the semantic information contained in the incomplete text (with mask tokens and error tokens) is missing too much. Thus, it is too difficult for FastCorrect to recover it. When the mask ratio is in a suitable range, FastCorrect will benefit from recovering the mask tokens. 2) the best choice of mask ratio is 0.15 for the FastCorrect.

\begin{table}[h]
\centering
\begin{tabular}{r|rr|rr}
\hline
\multicolumn{1}{l|}{}           & \multicolumn{2}{l|}{AISHELL-1}                      & \multicolumn{2}{l}{Inhouse Corpus}                 \\ \hline
\multicolumn{1}{l|}{Mask Ratio} & \multicolumn{1}{l}{WER} & \multicolumn{1}{l|}{WERR} & \multicolumn{1}{l}{WER} & \multicolumn{1}{l}{WERR} \\ \hline
0\%                               & 4.16                    & 13.87                     & 8.38                    & 6.79                     \\ \hline
10\%                             & 4.14                    & 15.75                     & 8.35                    & 7.12                     \\ \hline
\textbf{15\%}                            & \textbf{4.10}                    & \textbf{16.67}                     & \textbf{8.32}                    & \textbf{7.45}                     \\ \hline
20\%                             & 4.16                    & 15.30                     & 8.34                    & 7.23                     \\ \hline
25\%                            & 4.19                    & 14.61                     & 8.34                    & 7.23                     \\ \hline
30\%                             & 4.2                     & 14.38                     & 8.38                    & 6.79                     \\ \hline
40\%                             & 4.22                    & 13.93                     & 8.40                    & 6.56                     \\ \hline
50\%                             & 4.24                    & 13.47                     & 8.41                    & 6.45                     \\ \hline
\end{tabular}
\caption{The ablation experiments on the mask ratio hyper-parameter $p$ for ASR Error Correction. The best metrics are in bold. All the metrics are in \%.}
\label{table:asr_ablation_mask_ratio}
\end{table}

\section{Conclusion}
In this work, we proposed a effective and plug-and-play framework to address the trivial copy phenomenon, in which we randomly perturb a proportion of correct tokens into a special token \text{<mask>}. The model aims to predict the correct tokens from the special token. We apply this mechanism to both AR and NAR models for Spelling Error Correction tasks and the ASR Correction tasks. And experimental results show the effectiveness of the proposed method. Furthermore, our method can be easily extended to other tasks which suffer from heavy copy phenomenon.

\section{Acknowledgements}
This work has been supported in part by the Zhejiang NSF (LR21F020004), Chinese Knowledge Center of Engineering Science and Technology (CKCEST).

\section{Limitation}
In this work, we propose a framework MaskCorrect to address the heavily copy phenomenon in the spelling and ASR correction tasks. We partially solve the problem by a simple mask strategy with quantitative experiments. But, a more in-depth theoretical analysis of how to prevent the negative copy impact and how to adaptively mask the copied tokens is worth studying. Furthermore, despite the learning diagram we have explored, how to adapt the conventional model structure to cope with the heavily copied phenomenon is another future direction.

\bibliography{anthology,custom}
\bibliographystyle{acl_natbib}

\appendix

\section{Dataset Statistics}

\subsection{The Dataset Statistics for Spelling Error Correction}
\label{appendix:table-spelling_dataset_statistics}
We provide the detailed dataset statistics for spelling error correction in Table \ref{table-spelling_dataset_statistics}.
\begin{table}[h]
\centering
\begin{tabular}{l|ll}
\toprule
Traning data        & \# Sent   & Avg. Len \\ \midrule
Unpaired Corpus    & 5 million & 51       \\ \hline
Aug.                & 271329    & 44.4     \\
SIGHAN 14           & 6526      & 49.7     \\
SIGHAN 15           & 3174      & 30.0     \\
Paired Corpus Sum. & 281029    & 44.3     \\ \midrule \midrule
Test Data           & \# Sent   & Avg. Len \\ \hline
SIGHAN 14           & 1062      & 50.1     \\
SIGHAN 15           & 1100      & 30.5     \\ \hline
\end{tabular}
\caption{The dataset statistics for Spelling Error Correction.}
\label{table-spelling_dataset_statistics}
\end{table}

\subsection{The dataset statistics for ASR Error Correction}
\label{appendix:table-asr_dataset_statistics}
The AISHELL-1 dataset is a Mandarin Speech corpus containing 178 hours of training audios, 10 hours of validation audios and 5 hours of test audios. And the inhouse dataset is a much larger dataset for the industrial Mandarin ASR system. It contains 75K hours of Mandarin speech data for training and 200 hours of speech for both validation and test. 

\section{Dataset}

\subsection{Add Noise to unpaired data for Spelling Error Correction}
\label{appendix:add-noise_spelling}

We take the official confusion dictionary released by \cite{wu-etal-2013-chinese}\footnotemark[1], which is a collection of mapping from one Chinese character to its' widely mistaken characters. Following \cite{zhang-etal-2020-spelling}, we randomly replace 15\% of the characters in them with the mistaken characters by the confusion dictionary to artificially generate corresponding sentences with errors. 

\subsection{Add Noise to unpaired data for ASR Error Correction}
\label{appendix:add-noise_asr}
We first crawl 400M unpaired sentences from the Internet. Secondly, we find out the insertion, deletion, and substitution error rate in the training set. Finally, for each unpaired sentence, we apply random insertion, deletion, and substitution with a homophone dictionary based on the error rate statistics.

\subsection{The ASR Model for ASR Error Correction}
\label{appendix:asr_model_details}
As for the ASR model, we adopt a conformer with 12 layers encoder and 6 layers decoder, which is trained with cross-entropy loss and an auxiliary CTC loss on the encoder output. The hidden size of the conformer's encoder and decoder layers is 512. We use ESPnet\footnotemark[1] as the codebase. For the inhouse dataset, the ASR model is a hybrid model consisted of a latency-controlled BLSTM~\cite{zhang2016highway} acoustic model(am) and a 5-gram language model which is trained on 436 billion tokens.
\footnotetext[1]{\href{https://github.com/espnet/espnet}{https://github.com/espnet/espnet}}

\section{Implementation Details}
\subsection{Implementation Details of Spelling Error Correction}
\label{appendix:training_config_spell}
For a fair comparison, we choose the same pretrained BERT model for the above two methods provided by HuggingFace\footnotemark[2]. 
For the BERT correction model, we use 2 layers of MLPs with softmax to transform the bert embeddings to the target token probability distribution. Another 2 layers of MLPs are used to predict the error probability as an assistant loss. For the SoftMask BERT correction model, we use a bidirectional GRU with hidden dimension 768 as the error detection network. The MLP layers' dimension is set to 1024. The learning rate is set to 1e-4. We pretrain on the unpaired data with 9 epochs and finetune on the paired data with 4 epochs. The pretrain batch size is 1024 and the finetune batch size is 128. The mask probability of $m$ is 80\%, and the randomly replace probability of $n$ is 10\%. We mask the correct tokens with probability $p=$ 20\%. 
\footnotetext[2]{\href{https://huggingface.co/bert-base-chinese}{https://huggingface.co/bert-base-chinese}}

\subsection{Implementation Details of ASR Error Correction}
\label{appendix:training_config_asr}
We use SentencePiece~\cite{kudo2018sentencepiece} to learn the subwords and apply it to all the texts above. The dictionary size is set to 40K. We train all correction models on 4 NVIDIA Tesla V100 GPUs, with a batch size of 12000 tokens. The mask probability of $m$ is 80\%, and the randomly replace probability of $n$ is 10\%. We mask the correct tokens with probability $p=$ 15\%.
\begin{table*}[t]
\centering
\begin{tabular}{l|lll}
\toprule
Model       &  Precision & Recall & $F_{0.5}$ \\ \midrule
Transformer-big & 64.9 & 26.6 & 50.4 \\ \midrule
Transformer-big + MaskCorrect & \textbf{67.2} & \textbf{30.0} & \textbf{53.9} \\ \midrule
Sequence-to-Action & 65.9 &  28.9   &  52.5 \\ \bottomrule
\end{tabular}
\caption{The dataset statistics for Spelling Error Correction.}
\label{table:gec}
\end{table*}
\section{Experiments}
\subsection{Grammar Error Correction}
\label{sec:gec_exp}
In this section, we evaluate the effectiveness of our method on the Grammar Error Correction (GEC) task. 
\subsubsection{Datasets}
We conduct the experiments on the datasets of BEA-2019 GEC shared task~\cite{bryant2019bea}, restricted track. Specifically, the training set is consisted by four parts: Lang-8~\cite{mizumoto2011mining}, the FCE training set~\cite{yannakoudakis2011new}, NUCLE~\cite{dahlmeier2013building}, and \text{W\&I+LOCNESS}~\cite{granger2014computer,bryant2019bea}. We use the CoNLL-2013~\cite{ng-etal-2013-conll} test set as our dev set and the CoNLL-2014~\cite{ng2014conll} test set as our test set. We follow the S2A's~\cite{li2022sequence} practice and preprocess and tokenize all the sentences by 32K SentencePiece~\cite{kudo2018sentencepiece} Byte Pair Encoding (BPE)~\cite{sennrich2015neural}.

We use the official MaxMatch ($M^2$) scorer~\cite{dahlmeier2012better} to evaluate the models in the English GEC tasks. Given a source sentence, and a ground-truth sentence sample, the $M^2$ scorer searches for the highest overlap between them and reports the precision, recall, and $F_{0.5}$.

\subsubsection{Baseline Methods}
\paragraph{Transformer-big} The transformer~\cite{vaswani2017attention} architecture has shown its success in the sequence-to-sequence fields. We simply cast the GEC task as a sequence-to-sequence problem, which takes the sentences with errors as inputs and predicts the correct sentences autoregressively. We follow the default setting of transformer-big and implement it based on the official codes\footnotemark[1].
\footnotetext[1]{\href{https://github.com/facebookresearch/fairseq}{https://github.com/facebookresearch/fairseq}}.

\paragraph{Sequence-to-Action (S2A)} The sequence-to-action~\cite{li2022sequence} is a strong baseline based on sequence tagging, which is another diagram compared with seq2seq methods for GEC. The S2A method first predicts the token-level sequence actions and fuses the actions with the seq2seq framework to generate the final sentences.

\subsubsection{Implementation Details}
We evaluate our method on the transformer-big architecture. Similar to the previous tasks, we first collect the correct tokens from inputs and randomly mask the tokens with ratio $0.2$ and probability $m=0.8$ and $n=0.1$. We finetune the model by 2 epochs.

\subsubsection{Results Analysis}
In this section, we compare our method with the baseline models for the GEC task. The results are shown in Table \ref{table:gec}. From the results, we have several observations. Firstly, our method improves the transformer-big baseline for the English GEC task, which demonstrates the effectiveness of MaskCorrect. Secondly, with MaskCorrect, the transformer-big achieves state-of-art performance on all metrics.

\end{document}